\documentclass[10pt,conference]{IEEEtran}
\usepackage[table,x11names]{xcolor}
\usepackage{cite}
\usepackage{multirow}
\usepackage{hyperref}
\usepackage{amsfonts}
\usepackage[many]{tcolorbox}
\usepackage[tikz]{bclogo}
\usepackage{indentfirst}
\usepackage{enumitem}
\usepackage{xspace}
\usepackage{graphicx}
\usepackage{color}
\usepackage{float}
\usepackage{caption}
\usepackage{subcaption}

\definecolor{ghostwhite}{rgb}{0.86, 0.86, 0.86}

\newcommand{\etal}{\textit{et al}.\@\xspace}

\newtcolorbox{resultbox}[1][]{
  breakable,
  title=#1,
  colback=white,
  colbacktitle=white,
  coltitle=black,
  fonttitle=\bfseries,
  bottomrule=0pt,
  toprule=0pt,
  leftrule=3pt,
  rightrule=3pt,
  titlerule=0pt,
  arc=0pt,
  outer arc=0pt,
  colframe=black,
}

\begin{document}

\title{Supervised Sentiment Classification with CNNs for Diverse SE Datasets}

\author{
    \IEEEauthorblockN{Achyudh Ram, Meiyappan Nagappan}
    \IEEEauthorblockA{University of Waterloo
    \\\{arkeshav, mei.nagappan\}@uwaterloo.ca}
}

\maketitle

\begin{abstract}
Sentiment analysis, a popular technique for opinion mining, has been used by the software engineering research community for tasks such as assessing app reviews, developer emotions in issue trackers and developer opinions on APIs. Past research indicates that state-of-the-art sentiment analysis techniques have poor performance on SE data. This is because sentiment analysis tools are often designed to work on non-technical documents such as movie reviews. In this study, we attempt to solve the issues with existing sentiment analysis techniques for SE texts by proposing a hierarchical model based on convolutional neural networks (CNN) and long short-term memory (LSTM) trained on top of pre-trained word vectors. We assessed our model's performance and reliability by comparing it with a number of frequently used sentiment analysis tools on five gold standard datasets. Our results show that our model pushes the state of the art further on all datasets in terms of accuracy. We also show that it is possible to get better accuracy after labelling a small sample of the dataset and re-training our model rather than using an unsupervised classifier. 
\end{abstract}

\IEEEpeerreviewmaketitle

\section{\label{cha:intro}Introduction}

There are a number of studies that attempt to understand the role of sentiments in the software development process by classifying sentiments expressed by mobile users in app reviews, or by developers in issue trackers, code review tools and websites such as Stack Overflow \cite{guzman2014users, ortu2016emotional, ahmed2017senticr, pletea2014security}. From these studies, it could be seen that the effectiveness of sentiment analysis can be quite varied depending on the nature of the dataset used in the study and the task for which the tools used for performing the text classification were originally trained for. A lot of the existing studies use out-of-the-box sentiment analysis tools which were designed to work on non-technical documents such as movie reviews. This practice has been frequently criticized and has led to researchers developing tools specifically for software engineering related texts\cite{lin2018sentiment}. Tools frequently used by the SE community like SentiStrength and SentiStrengthSE were meant to be used on short texts \cite{thelwall2010sentistrength, islam2017leveraging}. However, common SE use-cases for these tools such as code reviews, discussions on issue trackers or on Stack Overflow often comprise of multiple long sentences. 

In this study, we address the issues raised by Lin \etal on using opinion mining in SE research \cite{lin2018sentiment}. Lin \etal state that there is no tool currently available for identifying sentiments expressed in SE related discussions and that re-training existing models on SE datasets does not improve the accuracy enough to justify expensive re-training for different datasets. The authors raise another concern that existing tools don't have acceptable precision and recall levels for tasks such as software library recommendations, and that blindly using the predicted sentiment would lead to wrong recommendations. We attempt to solve the issues with existing sentiment analysis techniques by proposing a hierarchical model based on convolutional neural networks (CNN) and long short-term memory (LSTM) networks trained on SE datasets that pushes the state of the art further. Even though convolutional neural networks were originally invented for analyzing visual imagery, existing studies have shown that they achieve state-of-the-art results for various sentence classification tasks, including single sentence sentiment prediction \cite{kim2014convolutional, zhang2015character}. LSTM is a popular neural network architecture that is composed of recurring units or cells that act as memory, thus enabling the network to learn long term dependencies. \cite{hochreiter1997lstm}. Since they have been designed to be able to     either retain or forget information in the cell state, these networks have been frequently used in studies and have achieved state-of-the-art results in various NLP tasks such as language modeling and machine translation \cite{sutskever2014sequence, jozefowicz2016exploring}. Zhou et al. show that an unified model that uses a CNN for feature extraction from phrases and an LSTM for obtaining the sentence representation outperforms a network that uses only a CNN or an LSTM \cite{zhou2015clstm}. In this study, we propose a unified hierarchical model in which we adapt the approach taken by Zhou et al. by having the CNN extract a sequence of representations for entire sentences rather than phrases, and the LSTM encode this sequence into a paragraph representation.

The goal of this study is to answer the following research questions:

\textbf{RQ1:} \textit{How does a unified hierarchical model perform when compared with other sentiment analysis tools on SE datasets?}  

\textbf{RQ2:} \textit{How does a unified hierarchical model scale with the amount of training data available when compared with other sentiment analysis tools?}

We assess our model's performance by comparing it with a number of frequently used sentiment analysis tools on five datasets that represent the SE scenarios in which researchers usually use these tools on. Even with just minimal tuning of hyperparameters, our simple model exceeds the current state of the art on all the five datasets. It also performs better on smaller datasets compared to other supervised classifiers. Further, it is considerably faster than the existing state of the art, Senti4SD. 

\section{\label{cha:related_work}Related work}

In this section, we look at the existing sentiment analysis tools frequently used by the SE community and relevant studies that compare the state of the art in sentiment analysis for SE datasets.

SentiStrength is one of the most widely adopted tools in the software engineering community for extracting sentiment strength from informal English text \cite{thelwall2010sentistrength}. SentiStrength outputs  both the positive and negative emotions for a sentence due to the fact that a sentence can have mixed sentiment. It was originally applied to social web texts but can be adjusted for other domains by adding new relevant words and sentiment strengths to the term list. SentiStrength-SE is built on top of the original SentiStrength for sentiment analysis specifically for the software engineering texts. Islam \etal showed that with heuristic improvements and a lexicon adjusted for technical texts, SentiStrengthSE outperforms SentiStrength on the JIRA issue comment benchmark dataset \cite{islam2017leveraging}. 

Senti4SD was developed by Calefato \etal for the specific purpose of performing sentiment analysis in a supervised setting on developer communication channels \cite{calefato2017sentiment}. A dataset of around 4,000 manually labelled questions, answers, and comments extracted from Stack Overflow was used for training and validating the classification algorithm. The authors claim that their classifier reduces the misclassifications of neutral and positive posts on their dataset when compared to SentiStrength. Senti4SD uses a rich feature space comprising of word embeddings, and n-gram, lexicon and keyword-based features.

SentiCR, a supervised sentiment analysis tool, was trained and validated on manually annotated code review comments from Gerrit \cite{ahmed2017senticr}. The tool is based on the Gradient Boosting Tree
(GBT) algorithm, and utilizes the bag of words model with Term
Frequency Inverse Document Frequency (TF-IDF) weights as features.
Apart from standard preprocessing, it performs synthetic minority over-sampling technique to address class imbalance in the dataset \cite{chawla2002smote}.  

VADER (Valence Aware Dictionary and sEntiment Reasoner) is a rule-based model for general sentiment analysis \cite{gilbert2014vader}. It is tuned for social media texts by incorporating an empirically validated sentiment lexicon with five rules that embody grammatical and syntactical conventions. The authors showed that VADER outperforms human raters and that it generalizes better than the other classifiers they used for comparison. 

From an analysis of the Apache Software Foundation issue tracker, Murgia \etal showed that development artifacts carry emotional information \cite{murgia2014developers} about the software development process. However, they state that greater the amount of context that is provided about an issue, greater is the extent to which human raters doubt their interpretation of emotions. A number of studies have benchmarked SE specific and off-the-shelf sentiment analysis toolkits on SE datasets: Jongeling \etal studied the extent of agreement of sentiments predicted by sentiment analysis tools such as SentiStrength, NLTK, Stanford NLP and Alchemy among themselves and with the sentiments recognized by human evaluators \cite{jongeling2015negative}; Novielli \etal perform a replication of this study to benchmark the performance of three SE specific sentiment analysis tools: SentiCR, Senti4SD and SentiStrengthSE \cite{novielli2018benchmark}; Lin \etal discuss the negative results obtained after training a recurrent neural network on a manually annotated dataset comprising of questions, answers and comments from
StackOverflow \cite{lin2018sentiment}. They further investigate the current state of the art by analyzing the performance of popular sentiment analysis toolkits on various SE datasets. The authors highlight the limitations of the sentiment analysis tools used by the SE community and state that efforts should be made to make sentiment analysis practical for SE research.

\section{\label{cha:background}Background}

\subsection{Sentiment analysis}
Sentiment analysis is the task of predicting the polarity of a phrase (such as positive or negative) often using natural language processing and machine learning techniques. A large part of sentiment analysis involves building predictive models that attempt to identify the emotional state (such as sadness or joy) or the nature of the opinion of a subject (such as positive or negative). Supervised techniques require labelled training data to train machine learning algorithms, whereas unsupervised methods can be applied in cases where labelling data is time consuming or expensive. Unsupervised classifiers use knowledge-based techniques or lexicon-based methods to perform the classification. Recently, there have been studies where supervised and unsupervised techniques are combined with a majority rule or voting classifier \cite{perea2013combining, medhat2014sentiment}. 

There are a number of commercial and open-source sentiment analysis tools available. In this study, we only consider tools that are publicly available and free for academic use. An overview of these tools along with the relevant research studies can be found in the next section. 

\subsection{Deep learning for sentence classification}
With rapidly increasing accessibility to deep learning, it has been applied to a wide range of NLP tasks, including sentence classification. These techniques capture contextual information better and mitigate the problems with the traditional bag-of-words model such as the curse of dimensionality\footnote{The curse of dimensionality refers to various phenomena that arise when analyzing and organizing data in high-dimensional spaces (often with hundreds or thousands of dimensions) that do not occur in low-dimensional settings. Source: \url{https://en.wikipedia.org/wiki/Curse_of_dimensionality}}. Two of the most widely used neural network architectures for natural language processing tasks are convolutional neural networks (CNN)
and recurrent neural networks (RNN). In this study, we use a variation of an RNN called the long short-term memory (LSTM).

\subsubsection{Convolutional Neural Networks (CNNs)}
A CNN typically comprises of multiple convolutional layers, each of which might be followed by a pooling layer, and finally a fully connected layer. A fully-connected neural network would not be practical in learning features from images due to the large number of neurons required for processing even relatively small images. A convolution layer, with the help of learn-able filters, provides a solution to this problem by reducing the number of parameters. Unlike a fully-connected neural network, each neuron is connected only to a local region of the input. Apart from a convolutional layer, these networks include pooling layers which essentially perform downsampling to reduce the spatial size of the representation. This reduces the number of parameters, thereby reducing the amount of computation required. Due to the fact that this architecture preserves the spatial structure of an image, it has been successfully used for computer vision tasks with minimal pre-processing required. 

Kim \etal showed that CNNs can achieve state-of-the-art results in single sentence sentiment prediction among other sentence classification tasks \cite{kim2014convolutional}. In this approach, the vector representations of the words in a sentence were concatenated vertically to create a two-dimensional matrix for each sentence. The resulting matrix was passed through a CNN to extract higher-level features for performing the classification. 

\subsubsection{Long-Short Term Memory (LSTM)}
Practical difficulties in training RNNs due to trade-offs between efficient learning and latching on to information for long periods have been observed and studied in detail \cite{bengio1994learning}.  LSTMs are a variation of RNNs designed to solve this long-term dependency problem \cite{hochreiter1997lstm}. Like RNNs, LSTMs have a chain like structure of repeating units. Unlike RNNs which have a single layer in the repeating units, LSTMs have a complex module of four interacting layers. The cell state aids in the flow of information across the chain of modules, and three gates control the addition or removal of information to the cell state. The ability of these networks to deal with variable-length input sequences and capture long-term dependencies have made these networks achieve remarkable results in a range of NLP tasks. 

A frequently used variant of RNNs called the Bidirectional-RNN was  proposed by Schuster \etal in 1997 \cite{schuster1997bidirectional}. In this architecture, classification is done on the combined outputs of two RNNs which process the input sentence from left to right and right to left. This approach can be extended to LSTM networks

\subsubsection{Word2Vec}
Word embedding is a language modelling technique used to create a continuous higher dimensional vector space representation of words such that similar words are closer to each other in that space. This learned distributed representation of words alleviates the curse of dimensionality \cite{bengio2003neural}. Word2Vec refers to a class of two layer neural network models that are trained on a large corpus of text to produce such word embeddings \cite{mikolov2013efficient}. There are two model architectures to learn the distributed representation of words:
\begin{enumerate}
	\item \textbf{Continuous bag-of-words  (CBOW):} In this architecture, the model predicts the current word from a window of surrounding context words. The order of context words does not influence the prediction \cite{mikolov2013efficient}.
	\item  \textbf{Skip-gram:} In this architecture, the model uses the current word to predict the surrounding window of context words. Unlike CBOW, the skip-gram architecture weighs nearby context words more heavily than more distant context words \cite{mikolov2013distributed}.
\end{enumerate}

According to Mikolov \etal,  the skip-gram architecture is slower but works better for infrequent words. In practice, a number of optimizations are made to increase the training speed such as the sub-sampling of high frequency words or negative sampling to ensure that each training sample only modifies the weights corresponding to a random selection of negative words. 

\section{\label{cha:methodology}Methodology}

\subsection{Model architecture}
\begin{figure*}
\includegraphics[width=0.75\textwidth]{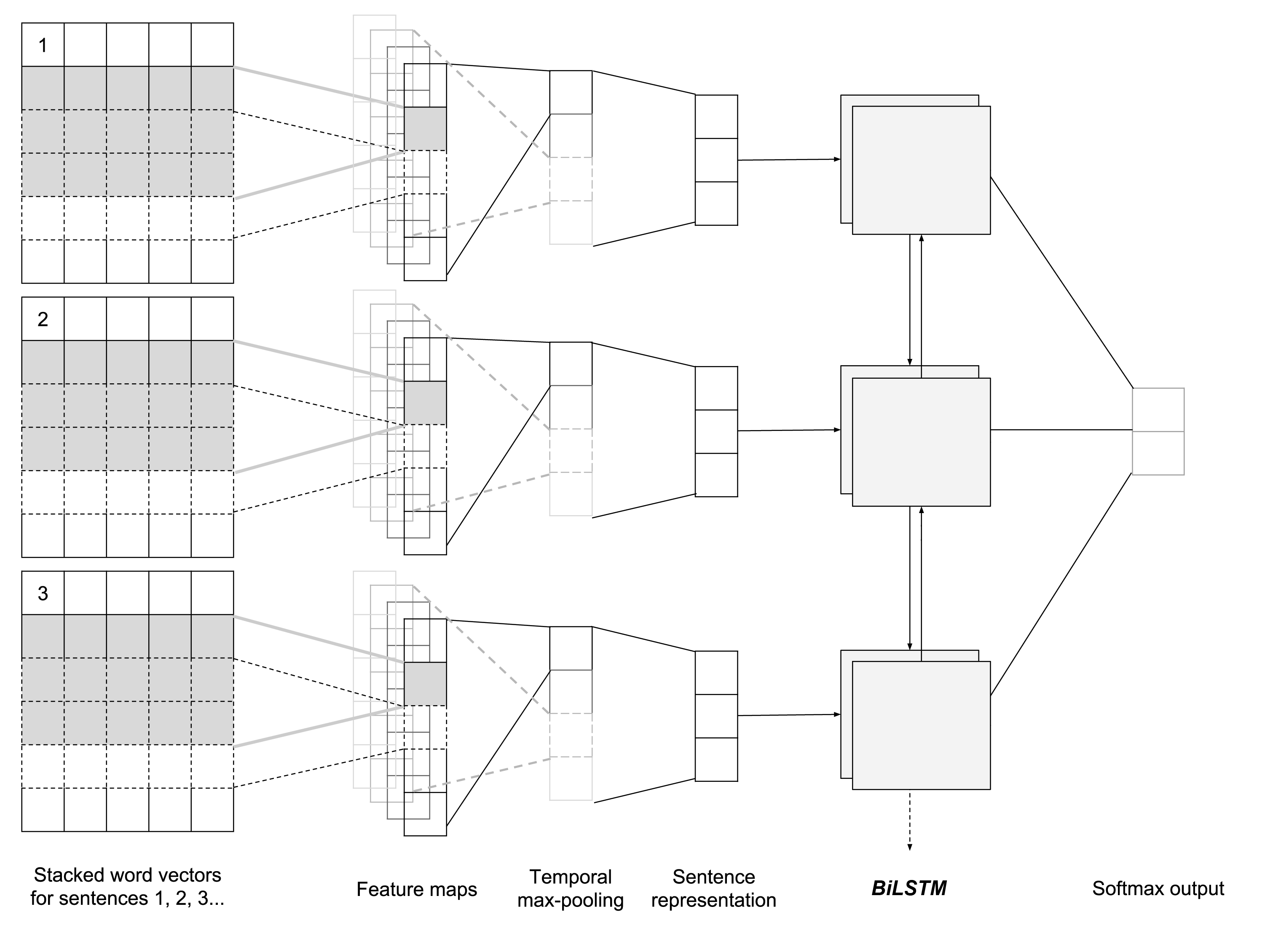}
\centering
\caption{Architecture of the unified hierarchical model}
\label{figure:architecture}
\end{figure*}

The architecture of the unified hierarchical model proposed in this study consists of 2 components: a CNN and an LSTM. The architecture of the proposed model can be seen in Figure \ref{figure:architecture}. In this section, we explain how the CNN is used to extract a sequence of representations for a sentence and how it works in unison with the LSTM to encode this sequence into a document/paragraph representation.

\subsubsection{Convolutional Neural Network (CNN)}
The CNN architecture used in this study (as shown in Figure \ref{figure:architecture}) is a variation of the single-channel architecture used by Kim et al. \cite{kim2014convolutional}. Let $n$ be the length of the longest sentence in the dataset. For a given sentence in the dataset, let $x_i$ be the k-dimensional word vector representing the $i^{th}$ word in that sentence. Every sentence $s$ in the dataset is represented as the concatenation (represented by the operator $\oplus$) of $x_i$, where $1 \leq i \leq n, i \in \mathbb{N}$:
\begin{equation} \label{eq:1}
s = x_1 \oplus x_2 \oplus ... \oplus x_{n-1} \oplus x_n
\end{equation}
Let us denote the concatenation of $(j+1)$ word vectors starting with $i$ as $x_{i:i+j}$. This resultant matrix is then passed through a temporal convolutional layer with a filter windows of size $f$ to produce new features. These filters are applied to each window of words in the sentence $s$ to produce a number of feature maps. A temporal max-pooling operation is applied to these feature maps to retain the feature with the highest value in every map. Finally, these features are fed to a fully connected layer of rectified linear units (ReLU) to create a $m$ dimensional vector representation of the sentence which is then passed on to an LSTM. 

\subsubsection{Long-Short Term Memory (LSTM)}
Every sample belonging to the dataset is treated as a document and is tokenized into sentences. Each sentence is passed through the CNN for extracting features as described above. For encoding these features, we use a Bidirectional-LSTM. The hidden states of the LSTM cells in the last time step is the encoded document/paragraph representation. This representation is passed to the final softmax layer for predicting the sentiment polarity, similar to the approach adopted by Zhou \etal \cite{zhou2015clstm}. 

\subsubsection{Regularization}
We employ dropout on the dense layers of our network for reducing overfitting \cite{srivastava2014dropout}. Dropout prevents complex co-adaptations of hidden units on training data by randomly removing (i.e. dropping out) hidden units along with their connections during training. We also employ dropout in the layers of the LSTM on the linear transformation of the inputs and the recurrent state.

\subsection{\label{sec:datasets}Datasets}
We evaluate our model on five publicly available SE benchmark datasets. Table \ref{tab:datasets} shows the size of the dataset, the number of samples and the class distribution.

The mobile app reviews dataset was created as a part of a study on the release planning of mobile apps by Villarroel \etal
 \cite{villarroel2016release}. Lin \etal randomly sampled and manually labelled 341 reviews from this dataset into positive, neutral and negative categories \cite{lin2018sentiment}. Two evaluators labelled the reviews, while a third resolved the conflicts. This dataset is considerably skewed with the minority (neutral) class consisting of only 25 samples. 

The Jira issue comments dataset by Ortu et al. \cite{ortu2015jira} has been used in various studies as one of the gold-standard dataset for sentiment analysis in software texts \cite{lin2018sentiment, novielli2018benchmark}. The dataset was labelled with six emotions: love, joy, anger, and sadness being the 4 most frequently expressed ones. As previously done by Lin et al. \cite{lin2018sentiment} and Jongeling et al. \cite{jongeling2015negative}, the issue comments labelled love or joy were considered as positive training samples, and those labelled anger or sadness were considered as negative training samples. The dataset has an imbalanced class distribution with 68.7\% of the issue comments belonging to the negative class. 

The Gerrit code review dataset was used for training and validating the SentiCR classification tool \cite{ahmed2017senticr}. It was annotated by three researchers, who classified review comments by following an ad-hoc approach into 3 classes: positive, neutral and negative. The former two were merged into one 'non-negative' category to reduce the class imbalance.

The Stack Overflow Java Libraries dataset was collected by Lin \etal for the purpose of recommending libraries based on sentiments mined from Stack Overflow \cite{lin2018sentiment}. It contains 1,500 randomly extracted sentences that have been labelled by five evaluators. This dataset is considerably skewed considering that it is labelled into 3 classes, with around 80\% of the samples belonging to the majority (neutral) class. 

The Stack Overflow Sentiments dataset was created as a part of a study to create a classifier for the analysis of sentiments in developer communication channels \cite{calefato2017sentiment}. This dataset, consisting of 4,423 questions, answers, and comments from Stack Overflow, was used for the training and validation of Senti4SD. Classes are relatively balanced in this dataset with 38\% of the posts belonging to the majority (neutral) class, and 27\% belonging to the minority (negative) class. This balance is the result of the authors performing sampling of posts based on the presence of affective lexicon as reflected by SentiStrength sentiment scores. 

\begin{table}[]
\centering
\caption{Gold-standard datasets used for evaluation}
\label{tab:datasets}
\begin{tabular}{|l|r|r|r|r|}
\hline
\multirow{2}{*}{Dataset} & \multicolumn{1}{c|}{\multirow{2}{*}{Number of samples}} & \multicolumn{3}{c|}{Class distribution} \\ \cline{3-5} & \multicolumn{1}{c|}{} & \multicolumn{1}{c|}{Positive} & \multicolumn{1}{c|}{Neutral} & \multicolumn{1}{c|}{Negative} \\ \hline
App Reviews & 341 & 54.5\% & 7.3\% & 38.2\% \\ \hline 
Jira & 926 & 31.3\% & N/A & 68.7\% \\ \hline
Gerrit & 1600 & 75.0\% & N/A & 25.0\% \\ \hline
SO Java Lib. & 1500 & 8.7\% & 79.4\% & 11.9\% \\ \hline
SO Sentiments & 4423 & 34.5\% & 38.3\% & 27.2\% \\ \hline
\end{tabular}
\end{table}

\subsection{Experimental setup}
The model was implemented with the Keras library\footnote{Available at \url{https://www.tensorflow.org/api_docs/python/tf/keras}} which comes packaged with the core Tensorflow API. This allows us to seamlessly shift between the TensorFlow and Keras workflows based on the task at hand. We train our model by minimizing the categorical cross-entropy. Stochastic gradient descent is used for training with the Adaptive Moment Estimation (Adam) optimizer  as it has been shown to work well in practice \cite{kingma2014adam}. We use early stopping when the validation loss has stopped decreasing to avoid overfitting when training \cite{prechelt1998automatic}. There are variations in the experimental setup across research questions. These variations are explained in Section \ref{cha:results}. The following six classifiers were used for comparing the performance of our model:

\begin{itemize}
	\item Supervised: Naive Bayes (NLTK), Senti4SD, SentiCR
	\item Unsupervised: SentiStrength, SentiStrength SE, VADER
\end{itemize}

\subsubsection{Hyperparameters}
We don't perform any dataset specific tuning of hyperparameters and hence the following hyperparameters apply for all the datasets. The temporal convolution layer has a filter window of size 5 with 150 feature maps each, and are activated by a rectified linear unit (ReLU). The dense layers have a dropout rate of 0.4. The layers of the LSTM have a dropout rate of 0.2 for both the inputs and the recurrent state.

\subsubsection{Pre-trained word vectors}
For all the datasets in this study, we use the 300-dimensional word vectors trained by Mikolov et al. on roughly 100 billion words from Google News\footnote{Available at https://code.google.com/archive/p/word2vec/}. We don't fine-tune the word vectors and keep them static throughout the analysis.

\subsubsection{Evaluation metrics}
For each of the models considered in the analysis, we follow the standard evaluation methodology by measuring the performance with the overall accuracy, along with precision, recall and F-score for each of the classes in the labelled dataset. Classification accuracy is the ratio of the number of correct predictions to the total samples in the dataset. For a given class, precision is the ratio of true positives to the number of predicted items of that class, and recall is the ratio of true positives to all the items that belong to that class. F-score is defined as the harmonic mean of precision and recall. Perfect precision and recall results in an F-score of one, and a zero precision or recall results in an F-score of zero. Precision, recall and F-score provide a complete picture of the model performance on datasets with unbalanced distribution of classes, whereas the overall accuracy allows a quick comparison of the performance of classifiers across all the classes.

\section{\label{cha:results}Results}

\subsection{Model performance evaluation}
\textbf{RQ1:} \textit{How does a unified hierarchical model perform when compared with other sentiment analysis tools on SE datasets?}  

To answer the first research question, we evaluate each classifier by performing a stratified 10-fold cross validation on the five datasets described in Section \ref{tab:datasets}. In stratified k-fold cross validation, each fold would contain approximately the same proportion of classes as the dataset, thereby making each fold a better representation of the dataset. We fixed the state of all random elements during the cross validation process so that each fold is identical for all the classifiers. The classification algorithm used for Senti4SD was L2 regularized logistic regression as it provided the best average performance across our datasets. For SentiCR, gradient boosted trees were used for classification based on the recommendation provided by the authors \cite{ahmed2017senticr}. The accuracy, precision, recall and F-measure for each classifier is reported in Table \ref{tab:rq1}. 

For every dataset, our model offers the best accuracy, along with consistently high precision and recall over all the polarity classes except the underrepresented classes of Stack Overflow Java Libraries and App Reviews datasets. Both of these datasets have a sharply skewed distribution of classes, and the relatively small number of samples is not sufficient for training any of the supervised classifiers. Even the unsupervised classifiers don't offer satisfactory performance in this regard.

\begin{table*}
\centering
{\footnotesize
\caption{Performance of classifiers on gold-standard datasets}
\label{tab:rq1}
\begin{tabular}{|l|l|r|r|r|r|r|r|r|r|r|r|}
\hline
\multirow{2}{*}{Dataset} & \multirow{2}{*}{Classifier} & \multicolumn{3}{c|}{Negative} & \multicolumn{3}{c|}{Positive} & \multicolumn{3}{c|}{Neutral} & \multicolumn{1}{c|}{\multirow{2}{*}{Acc.}} \\ \cline{3-11}
 &  & \multicolumn{1}{c|}{P} & \multicolumn{1}{c|}{R} & \multicolumn{1}{c|}{F1} & \multicolumn{1}{c|}{P} & \multicolumn{1}{c|}{R} & \multicolumn{1}{c|}{F1} & \multicolumn{1}{c|}{P} & \multicolumn{1}{c|}{R} & \multicolumn{1}{c|}{F1} & \multicolumn{1}{c|}{} \\ \hline
 
\multirow{7}{*}{Jira} & Naive Bayes & 0.94 & 0.94 & 0.94 & 0.88 & 0.86 & 0.87 & - & - & - & 0.92 \\ \cline{2-12} 
 & VADER & \textbf{0.99} & 0.72 & 0.83 & 0.62 & \textbf{0.99} & 0.76 & - & - & - & 0.80 \\ \cline{2-12} 
 & SentiStrength & \textbf{0.99} & 0.70 & 0.82 & 0.60 & \textbf{0.99} & 0.75 & - & - & - & 0.79 \\ \cline{2-12} 
 & SentiStrengthSE & \textbf{0.99} & 0.70 & 0.82 & 0.61 & \textbf{0.99} & 0.75 & - & - & - & 0.80 \\ \cline{2-12} 
 & Senti4SD & 0.86 & 0.96 & 0.91 & 0.93 & 0.92 & 0.92 & - & - & - & 0.95 \\ \cline{2-12} 
 & SentiCR & 0.94 & \textbf{0.99} & 0.96 & 0.97 & 0.86 & 0.91 & - & - & - & 0.95 \\ \cline{2-12} \rowcolor{ghostwhite}
 & Hi-CNN-LSTM & 0.97 & \textbf{0.99} & \textbf{0.98} & \textbf{0.98} & 0.92 & \textbf{0.95} & - & - & - & \textbf{0.97} \\ \hline
 
\multirow{7}{*}{\begin{tabular}[c]{@{}l@{}}App \\ Reviews\end{tabular}} & Naive Bayes & 0.88 & 0.65 & 0.75 & 0.78 & 0.84 & 0.81 & 0.18 & 0.30 & 0.22 & 0.72 \\ \cline{2-12} 
 & VADER & \multicolumn{1}{l|}{0.87} & \multicolumn{1}{l|}{0.44} & \multicolumn{1}{l|}{0.58} & \multicolumn{1}{l|}{0.66} & \multicolumn{1}{l|}{0.92} & \multicolumn{1}{l|}{0.77} & \multicolumn{1}{l|}{\textbf{0.21}} & \multicolumn{1}{l|}{0.16} & \multicolumn{1}{l|}{\textbf{0.18}} & \multicolumn{1}{l|}{0.68} \\ \cline{2-12} 
 & SentiStrength & 0.81 & 0.34 & 0.48 & 0.74 & 0.86 & 0.80 & 0.11 & 0.32 & 0.16 & 0.62 \\ \cline{2-12} 
 & SentiStrengthSE & \textbf{0.93} & 0.30 & 0.45 & 0.72 & 0.74 & 0.73 & 0.09 & \textbf{0.40} & 0.15 & 0.54 \\ \cline{2-12} 
 & Senti4SD & 0.77 & 0.81 & 0.79 & 0.84 & 0.90 & 0.87 & 0.10 & 0.05 & 0.07 & 0.81 \\ \cline{2-12} 
 & SentiCR & 0.82 & 0.75 & 0.78 & 0.83 & 0.91 & 0.87 & 0.12 & 0.13 & 0.12 & 0.79 \\ \cline{2-12} \rowcolor{ghostwhite}
 & Hi-CNN-LSTM & 0.86 & \textbf{0.87} & \textbf{0.86} & \textbf{0.85} & \textbf{0.94} & \textbf{0.89} & 0.15 & 0.08 & 0.10 & \textbf{0.85} \\ \hline
 
\multirow{7}{*}{Gerrit} & Naive Bayes & 0.52 & 0.43 & 0.47 & 0.82 & 0.86 & 0.84 & - & - & - & 0.75 \\ \cline{2-12} 
 & VADER & 0.43 & 0.45 & 0.44 & 0.81 & 0.80 & 0.80 & - & - & - & 0.72 \\ \cline{2-12} 
 & SentiStrength & \textbf{0.99} & \textbf{0.70} & \textbf{0.82} & 0.60 & \textbf{0.99} & 0.75 & - & - & - & 0.78 \\ \cline{2-12} 
 & SentiStrengthSE & 0.54 & 0.30 & 0.39 & 0.80 & 0.92 & 0.86 & - & - & - & 0.76 \\ \cline{2-12} 
 & Senti4SD & 0.75 & 0.49 & 0.59 & 0.85 & 0.93 & \textbf{0.89} & - & - & - & 0.82 \\ \cline{2-12} 
 & SentiCR & 0.61 & 0.66 & 0.63 & \textbf{0.88} & 0.86 & 0.87 & - & - & - & 0.81 \\ \cline{2-12} \rowcolor{ghostwhite}
 & Hi-CNN-LSTM & 0.75 & 0.46 & 0.57 & 0.84 & 0.95 & \textbf{0.89} & - & - & - & \textbf{0.83} \\ \hline
 
\multirow{7}{*}{\begin{tabular}[c]{@{}l@{}}SO\\ Sentiments\end{tabular}} & Naive Bayes & 0.59 & 0.57 & 0.58 & 0.87 & 0.75 & 0.81 & 0.63 & 0.73 & 0.67 & 0.69 \\ \cline{2-12} 
 & VADER & 0.67 & 0.79 & 0.73 & 0.69 & \textbf{0.94} & 0.80 & 0.85 & 0.47 & 0.61 & 0.72 \\ \cline{2-12} 
 & SentiStrength & 0.67 & \textbf{0.93} & 0.78 & 0.89 & 0.92 & 0.90 & \textbf{0.92} & 0.63 & 0.75 & 0.81 \\ \cline{2-12} 
 & SentiStrengthSE & 0.75 & 0.76 & 0.75 & \textbf{0.91} & 0.82 & 0.86 & 0.72 & 0.79 & 0.75 & 0.79 \\ \cline{2-12} 
 & Senti4SD & 0.79 & 0.85 & \textbf{0.82} & 0.90 & 0.92 & \textbf{0.91} & 0.84 & 0.79 & 0.81 & 0.85 \\ \cline{2-12} 
 & SentiCR & 0.80 & 0.73 & 0.76 & 0.89 & 0.91 & 0.90 & 0.79 & 0.82 & 0.80 & 0.83 \\ \cline{2-12} \rowcolor{ghostwhite}
 & Hi-CNN-LSTM & \textbf{0.84} & 0.81 & \textbf{0.82} & 0.90 & 0.92 & \textbf{0.91} & 0.83 & \textbf{0.83} & \textbf{0.83} & \textbf{0.86} \\ \hline
 
\multirow{7}{*}{\begin{tabular}[c]{@{}l@{}}SO\\ Java Lib.\end{tabular}} & Naive Bayes & 0.54 & 0.38 & 0.45 & 0.46 & 0.22 & 0.30 & 0.85 & 0.93 & 0.89 & 0.80 \\ \cline{2-12} 
 & VADER & 0.47 & 0.51 & 0.49 & 0.19 & \textbf{0.64} & 0.29 & 0.89 & 0.74 & 0.81 & 0.63 \\ \cline{2-12} 
 & SentiStrength & 0.39 & 0.43 & 0.41 & 0.20 & 0.36 & 0.26 & 0.86 & 0.76 & 0.81 & 0.69 \\ \cline{2-12} 
 & SentiStrengthSE & 0.50 & 0.18 & 0.26 & 0.31 & 0.22 & 0.26 & 0.82 & 0.93 & 0.87 & 0.78 \\ \cline{2-12} 
 & Senti4SD & \textbf{0.55} & 0.35 & 0.43 & \textbf{0.65} & 0.16 & 0.26 & 0.85 & 0.96 & \textbf{0.90} & \textbf{0.82} \\ \cline{2-12} 
 & SentiCR & 0.50 & \textbf{0.67} & \textbf{0.57} & 0.48 & 0.36 & \textbf{0.41} & \textbf{0.90} & 0.87 & 0.88 & 0.80 \\ \cline{2-12} \rowcolor{ghostwhite}
 & Hi-CNN-LSTM & 0.40 & 0.22 & 0.28 & 0.21 & 0.07 & 0.11 & 0.83 & \textbf{0.99} & \textbf{0.90} & \textbf{0.82} \\ \hline
\end{tabular}}
\end{table*}

\subsection{Model scalability with dataset size}

\textbf{RQ2:} \textit{How does a unified hierarchical model scale with the amount of training data available when compared with other sentiment analysis tools?}

The main goal of this research question was to observe the change in validation accuracy with the amount of data available to our model, thereby identifying the amount of data required to make the most of our model. We also wanted to identify the lowest amount of labelling to be done by researchers in order for a supervised classifier to exceed the performance of a state-of-the-art unsupervised classifier. Answering this question allows us to provide a ballpark dataset size below which it would be better to use an unsupervised algorithm rather than spend time on labelling. We also compare how the performance of other classifiers evolve with the amount of data available.

\begin{figure*}
    \centering
    \begin{subfigure}[b]{0.45\textwidth}
        \centering
        \includegraphics[width=\textwidth]{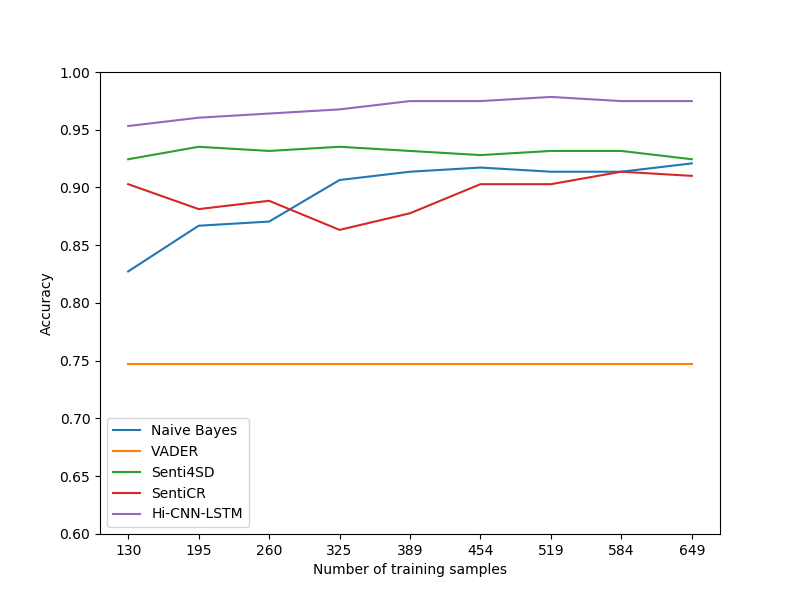}
        \caption{Jira}
    \end{subfigure}
	~
    \begin{subfigure}[b]{0.45\textwidth}
        \centering
        \includegraphics[width=\textwidth]{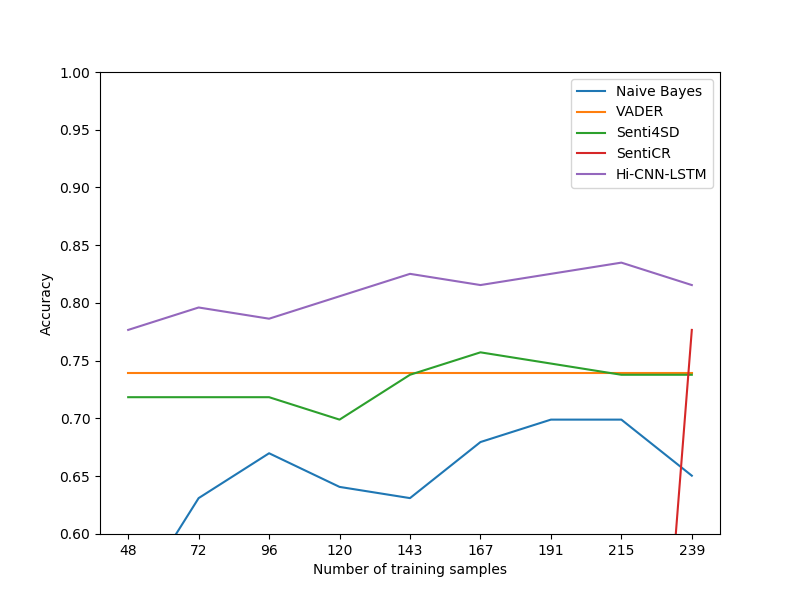}
        \caption{App Reviews}
    \end{subfigure}
    	~
    \begin{subfigure}[b]{0.45\textwidth}
        \centering
        \includegraphics[width=\textwidth]{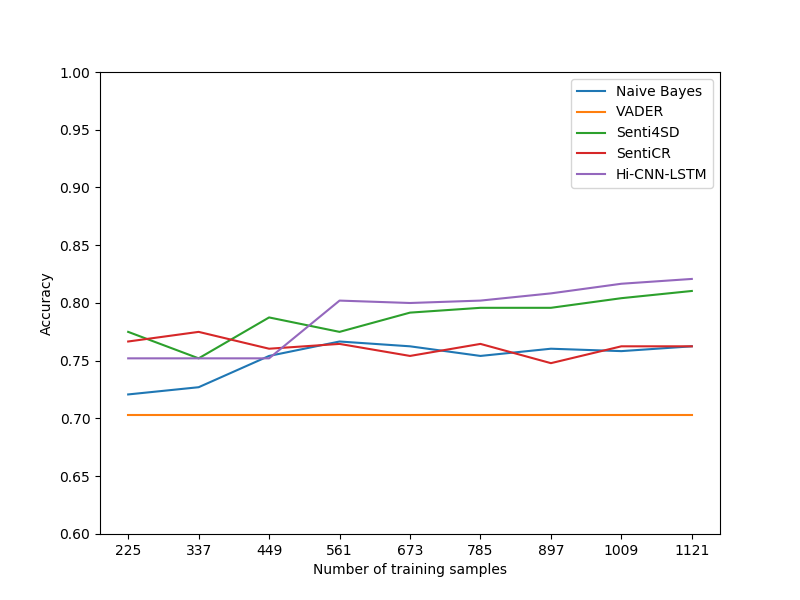}
        \caption{Gerrit}
    \end{subfigure}
    ~
    \begin{subfigure}[b]{0.45\textwidth}
        \centering
        \includegraphics[width=\textwidth]{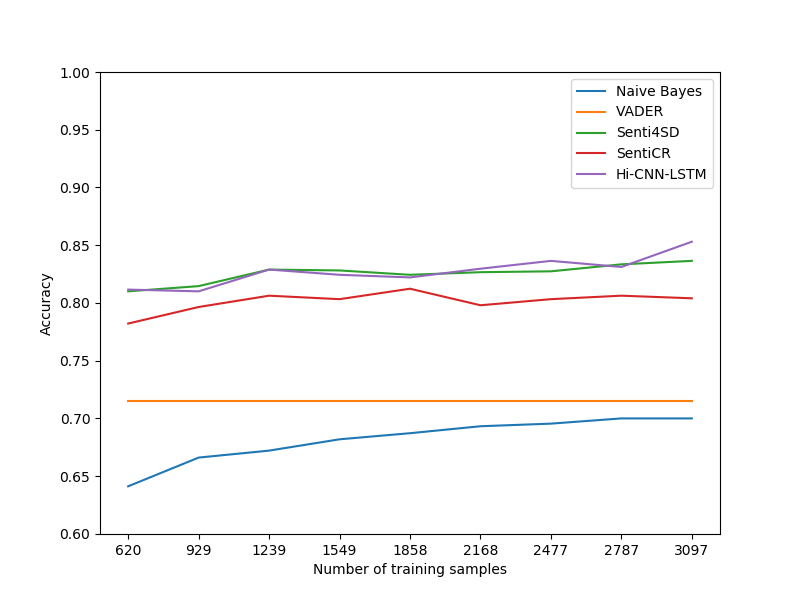}
        \caption{Stack Overflow Sentiments}
    \end{subfigure}
    ~
    \begin{subfigure}[b]{0.45\textwidth}
        \centering
        \includegraphics[width=\textwidth]{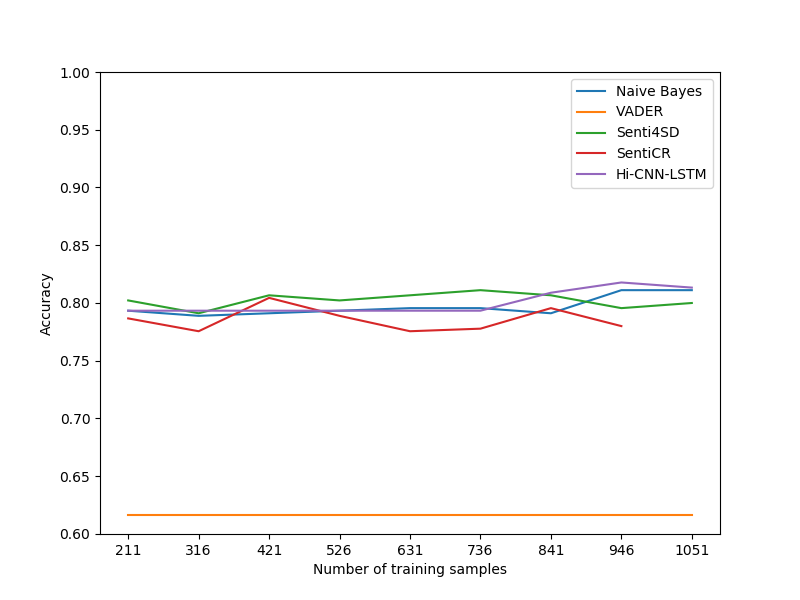}
        \caption{Stack Overflow Java Lib.}
    \end{subfigure}
    
    \caption{Variation in accuracy with the amount of training data}
    \label{figure:trend_graph}
\end{figure*}

To answer this research question, we performed a test-train split of 70-30 percent. We then resampled the elements of the train dataset with replacement with increasing sizes starting at 20\% of the train dataset size.  We fixed the state of all random elements during the test-train split and resampling process in order to make a fair comparison across classifiers. We were unable to train SentiCR on the resampled splits of the App Reviews dataset due to the small size of the dataset. We present the results of our analysis in Figure \ref{figure:trend_graph}. We denote the performance of VADER, an unsupervised classifier, on the test split by a horizontal line.  

The results from Figure \ref{figure:trend_graph} show that, depending on the dataset, having even a small number of labels would help train a supervised classifier that performs better than an unsupervised classifier. Naive Bayes, the baseline classifier, is the only supervised classifier to perform worse than VADER. This makes a case for labelling a small balanced dataset and trying out supervised classification algorithms before resorting to unsupervised tools. Depending on the results, more labelling can be done to improve the results further. 

Even though the accuracy rises with the number of training samples available, the increase is very gradual. In a bi-polar dataset that is easy to classify such as Jira, quintupling the size of the training data from 130 to 649 leads to an increase in accuracy of only 2\%. Even though most supervised tools struggle to outperform VADER on the App Reviews dataset due to its small size, our hierarchical model is more accurate than VADER after training on just 48 samples.

\section{\label{cha:title}Discussion}

From the above results, we can understand that deep learning models have a lot to offer when it comes to building customized sentiment analysis tools for the SE community. We also make a case for why supervised techniques are better for sentiment analysis tasks despite requiring labelled datasets. Using state of the art deep learning techniques, we were able to build a classifier that not only performs better than the existing tools, but also scales better when there aren't a lot of training samples. Further, our model opens up the possibility of transfer learning by using our trained model for feature extraction. 

One of the motivations behind developing this model was that the sentiment analysis tools frequently used by the SE community are not optimized for long texts. This often necessitated researchers to label individual lines rather than a block of text like an entire comment or a bug report. Some of the datasets used in this study such as App Reviews, Gerrit and SO Sentiment contain labels for multiple sentences and our model outperforms all of the existing tools on these datasets. However, it is not possible to conclusively measure the performance improvement obtained when using our classifier on long texts without performing an ablation study.

One of the concerns raised by Lin \etal on re-training existing models on SE datasets is that there isn't enough improvement in accuracy to justify the expensive and time consuming training process for each dataset \cite{lin2018sentiment}. While validating the performance of other supervised classifiers, we also noticed the considerably longer time required to train and test Senti4SD. Table \ref{tab:rq1} showed that our tool makes the most accurate predictions on all of the five datasets, followed by Senti4SD. We measured the time taken for training and inference during one fold of a 10-fold cross validation to compare the time performance of both of these models. Table \ref{tab:time} shows the time taken on a PC equipped with Intel Core i7-6770HQ and 16 GB of RAM. These values take into account the time taken for feature extraction and parameter tuning for Senti4SD. It can be seen that our model is on average twice as fast for training and 200 times faster for inference compared to Senti4SD despite providing equivalent or better predictions on all of these datasets. 

Since we employ validation-based early stopping, training is stopped when there isn't a steady decline in the validation loss. Since the network trains until it is stopped, the time taken for training is not solely determined by the size of the training set, but also on factors like the number of epochs of training that can be run before the network begins to overfit. Despite the Gerrit dataset being smaller than the Stack Overflow Sentiments dataset (around one-third the size), our model takes longer to train on the Gerrit dataset. For a given model, the time taken for inference, however, is largely dependent on the size of the test set. It should be noted that our model scales well on GPUs as it is based on the TensorFlow deep learning framework. With a mid-range GPU such as the Nvidia GTX 1050, we observed that our model ran more than 6 times faster than the numbers reported in Table \ref{tab:time}. 

\begin{table}[]
\centering
\caption{Time (in secs.) for training and testing of one fold in a 10-fold cross validation}
\label{tab:time}
\begin{tabular}{|l|r|r|r|r|}
\hline
\multirow{2}{*}{Dataset} & \multicolumn{2}{c|}{Senti4SD}                                                & \multicolumn{2}{l|}{Hi-CNN-LSTM}                                 \\ \cline{2-5} 
                         & \multicolumn{1}{l|}{Train} & \multicolumn{1}{l|}{Test} & \multicolumn{1}{l|}{Train} & \multicolumn{1}{l|}{Test} \\ \hline
Jira & 868.54 & 135.43 & 29.87 & 0.32 \\ \hline
App Reviews & 387.78 & 80.71 & 22.20 & 0.36 \\ \hline
Gerrit & 1443.03 & 199.68 & 359.02 & 1.34 \\ \hline
SO Sentiments & 3839.34 & 445.67 & 430.63 & 2.13 \\ \hline
SO Java Lib. & 1340.19 & 183.91 & 13.33 & 0.52 \\ \hline
\end{tabular}
\end{table}

\section{\label{cha:threats}Threats to Validity}

\subsection{Internal validity}
Wherever we could, we used hyperparameters and dataset pre-processing methods recommended by the authors while training the classifiers from other studies. However, since we did not conduct an exhaustive grid search for optimizing hyperparameters on these classifiers due to time and resource constraints, we could be reporting sub-optimal performance metrics \cite{fu2016differential}. 


\subsection{External validity}
Sentiment analysis datasets in SE can be quite diverse based on the application. This makes it hard to make a model that generalizes universally across datasets even within the SE domain. In order to ensure that our model does not overfit, we employ dropout on the fully-connected and recurrent layers, along with validation-based early stopping. However, more datasets will be required to ascertain how well our model generalizes.

It is hard to explain the positive or negative results associated with deep learning as the cause of the result could be impossible to pinpoint. The explanations given regarding the choice of hyperparameters or the network architecture is purely based on our intuition and the empirical results from existing literature. This also makes it hard to comment on whether the hyperparameters and the network architecture chosen based on the datasets in this study would generalize to other datasets in the future. In order to minimize this threat, we chose five datasets on sentiment analysis from diverse SE domains. 

\section{\label{cha:conc}Conclusion}

In this study, we described a novel unified hierarchical model benchmarked on five SE sentiment analysis datasets. Our results show that this model provides more accurate predictions than the existing state-of-the-art supervised and unsupervised classifiers, despite the lack of extensive hyperparameter tuning. Further, it scales better to small datasets compared to other supervised classifiers, and takes less time to train. We also discuss why it is better to label a small sample of the dataset and train a supervised classifier rather than using an unsupervised classifiers.

We initially set out to address the issues raised by Lin \etal \cite{lin2018sentiment} on opinion mining in SE research and show that by either finding appropriate existing models and retraining them on the new dataset or developing models specifically for the SE application, it is possible to achieve satisfactory results that are usable for research purposes. When we identified existing datasets and validated the performance of existing classifiers, we noticed that Senti4SD and SentiCR provide usable results on most of the datasets considered, with minor exceptions. All of the supervised classifiers considered in this study perform poorly on the minority classes of the App Reviews and SO Java Lib. datasets due to the lack of enough training samples for these classes. While Lin \etal discuss that the nature of the data in certain SE applications (such as issue trackers) might make sentiment analysis easier, they don't comprehensively consider a wide range of SE datasets. Of the three datasets they look at (App Reviews, Jira, and SO Java Lib.), two of them have a skewed class distribution. Further, the authors didn't consider supervised tools built specifically for SE such as Senti4SD and SentiCR. In this paper, we would like to make a case that opinion mining is very well feasible for SE research with the caveat that it is at least required to re-train existing models on the appropriate domain data, and in some cases building models and tweaking hyperparameters based on the nature of the data. 

The primary contributions of this study are:
\begin{enumerate}
	\item A novel supervised sentiment classification model that provides state of the art performance on a diverse range of SE datasets. The classifier and the datasets used in this study are publicly available at \url{https://github.com/achyudhk/SentiGH}.
	\item A comprehensive comparative analysis of existing sentiment analysis tools and how they scale with the amount of available training data.
\end{enumerate}

\bibliographystyle{IEEEtran}
\bibliography{bare_conf}

\begin{thebibliography}{10}
\providecommand{\url}[1]{#1}
\csname url@samestyle\endcsname
\providecommand{\newblock}{\relax}
\providecommand{\bibinfo}[2]{#2}
\providecommand{\BIBentrySTDinterwordspacing}{\spaceskip=0pt\relax}
\providecommand{\BIBentryALTinterwordstretchfactor}{4}
\providecommand{\BIBentryALTinterwordspacing}{\spaceskip=\fontdimen2\font plus
\BIBentryALTinterwordstretchfactor\fontdimen3\font minus
  \fontdimen4\font\relax}
\providecommand{\BIBforeignlanguage}[2]{{%
\expandafter\ifx\csname l@#1\endcsname\relax
\typeout{** WARNING: IEEEtran.bst: No hyphenation pattern has been}%
\typeout{** loaded for the language `#1'. Using the pattern for}%
\typeout{** the default language instead.}%
\else
\language=\csname l@#1\endcsname
\fi
#2}}
\providecommand{\BIBdecl}{\relax}
\BIBdecl

\bibitem{guzman2014users}
E.~Guzman and W.~Maalej, ``How do users like this feature? a fine grained
  sentiment analysis of app reviews,'' in \emph{Requirements Engineering
  Conference (RE), 2014 IEEE 22nd International}.\hskip 1em plus 0.5em minus
  0.4em\relax IEEE, 2014, pp. 153--162.

\bibitem{ortu2016emotional}
M.~Ortu, A.~Murgia, G.~Destefanis, P.~Tourani, R.~Tonelli, M.~Marchesi, and
  B.~Adams, ``The emotional side of software developers in jira,'' in
  \emph{Proceedings of the 13th International Conference on Mining Software
  Repositories}.\hskip 1em plus 0.5em minus 0.4em\relax ACM, 2016, pp.
  480--483.

\bibitem{ahmed2017senticr}
T.~Ahmed, A.~Bosu, A.~Iqbal, and S.~Rahimi, ``{SentiCR: A Customized Sentiment
  Analysis Tool for Code Review Interactions},'' in \emph{32nd IEEE/ACM
  International Conference on Automated Software Engineering (NIER track)},
  ser. ASE '17, 2017.

\bibitem{pletea2014security}
D.~Pletea, B.~Vasilescu, and A.~Serebrenik, ``Security and emotion: sentiment
  analysis of security discussions on github,'' in \emph{Proceedings of the
  11th working conference on mining software repositories}.\hskip 1em plus
  0.5em minus 0.4em\relax ACM, 2014, pp. 348--351.

\bibitem{lin2018sentiment}
B.~Lin, F.~Zampetti, G.~Bavota, M.~Di~Penta, M.~Lanza, and R.~Oliveto,
  ``Sentiment analysis for software engineering: How far can we go?'' in
  \emph{IEEE/ACM 40th International Conference on Software Engineering (ICSE)},
  2018.

\bibitem{thelwall2010sentistrength}
\BIBentryALTinterwordspacing
M.~Thelwall, K.~Buckley, G.~Paltoglou, D.~Cai, and A.~Kappas, ``Sentiment in
  short strength detection informal text,'' \emph{J. Am. Soc. Inf. Sci.
  Technol.}, vol.~61, no.~12, pp. 2544--2558, Dec. 2010. [Online]. Available:
  \url{http://dx.doi.org/10.1002/asi.v61:12}
\BIBentrySTDinterwordspacing

\bibitem{islam2017leveraging}
M.~R. Islam and M.~F. Zibran, ``Leveraging automated sentiment analysis in
  software engineering,'' in \emph{Proceedings of the 14th International
  Conference on Mining Software Repositories}.\hskip 1em plus 0.5em minus
  0.4em\relax IEEE Press, 2017, pp. 203--214.

\bibitem{kim2014convolutional}
Y.~Kim, ``Convolutional neural networks for sentence classification,'' in
  \emph{Proceedings of the 2014 Conference on Empirical Methods in Natural
  Language Processing (EMNLP)}, 2014, pp. 1746--1751.

\bibitem{zhang2015character}
X.~Zhang, J.~Zhao, and Y.~LeCun, ``Character-level convolutional networks for
  text classification,'' in \emph{Advances in neural information processing
  systems}, 2015, pp. 649--657.

\bibitem{hochreiter1997lstm}
S.~Hochreiter and J.~Schmidhuber, ``Long short-term memory,'' \emph{Neural
  computation}, vol.~9, no.~8, pp. 1735--1780, 1997.

\bibitem{sutskever2014sequence}
I.~Sutskever, O.~Vinyals, and Q.~V. Le, ``Sequence to sequence learning with
  neural networks,'' in \emph{Advances in neural information processing
  systems}, 2014, pp. 3104--3112.

\bibitem{jozefowicz2016exploring}
R.~Jozefowicz, O.~Vinyals, M.~Schuster, N.~Shazeer, and Y.~Wu, ``Exploring the
  limits of language modeling,'' \emph{arXiv preprint arXiv:1602.02410}, 2016.

\bibitem{zhou2015clstm}
C.~Zhou, C.~Sun, Z.~Liu, and F.~Lau, ``A c-lstm neural network for text
  classification,'' \emph{arXiv preprint arXiv:1511.08630}, 2015.

\bibitem{calefato2017sentiment}
F.~Calefato, F.~Lanubile, F.~Maiorano, and N.~Novielli, ``Sentiment polarity
  detection for software development,'' \emph{Empirical Software Engineering},
  pp. 1--31, 2017.

\bibitem{chawla2002smote}
N.~V. Chawla, K.~W. Bowyer, L.~O. Hall, and W.~P. Kegelmeyer, ``Smote:
  synthetic minority over-sampling technique,'' \emph{Journal of artificial
  intelligence research}, vol.~16, pp. 321--357, 2002.

\bibitem{gilbert2014vader}
C.~H.~E. Gilbert, ``Vader: A parsimonious rule-based model for sentiment
  analysis of social media text,'' 2014.

\bibitem{murgia2014developers}
A.~Murgia, P.~Tourani, B.~Adams, and M.~Ortu, ``Do developers feel emotions? an
  exploratory analysis of emotions in software artifacts,'' in
  \emph{Proceedings of the 11th working conference on mining software
  repositories}.\hskip 1em plus 0.5em minus 0.4em\relax ACM, 2014, pp.
  262--271.

\bibitem{jongeling2015negative}
Jongeling, Datta, and Serebrenik, ``Choosing your weapons: On sentiment
  analysis tools for software engineering research,'' in \emph{2015 IEEE
  International Conference on Software Maintenance and Evolution (ICSME)}, Sept
  2015, pp. 531--535.

\bibitem{novielli2018benchmark}
N.~Novielli, D.~Girardi, and F.~Lanubile, ``A benchmark study on sentiment
  analysis for software engineering research,'' \emph{arXiv preprint
  arXiv:1803.06525}, 2018.

\bibitem{perea2013combining}
J.~M. Perea-Ortega, E.~Mart{\'\i}nez-C{\'a}mara, M.-T. Mart{\'\i}n-Valdivia,
  and L.~A. Ure{\~n}a-L{\'o}pez, ``Combining supervised and unsupervised
  polarity classification for non-english reviews,'' in \emph{International
  Conference on Intelligent Text Processing and Computational
  Linguistics}.\hskip 1em plus 0.5em minus 0.4em\relax Springer, 2013, pp.
  63--74.

\bibitem{medhat2014sentiment}
W.~Medhat, A.~Hassan, and H.~Korashy, ``Sentiment analysis algorithms and
  applications: A survey,'' \emph{Ain Shams Engineering Journal}, vol.~5,
  no.~4, pp. 1093--1113, 2014.

\bibitem{bengio1994learning}
Y.~Bengio, P.~Simard, and P.~Frasconi, ``Learning long-term dependencies with
  gradient descent is difficult,'' \emph{IEEE transactions on neural networks},
  vol.~5, no.~2, pp. 157--166, 1994.

\bibitem{schuster1997bidirectional}
M.~Schuster and K.~K. Paliwal, ``Bidirectional recurrent neural networks,''
  \emph{IEEE Transactions on Signal Processing}, vol.~45, no.~11, pp.
  2673--2681, 1997.

\bibitem{bengio2003neural}
Y.~Bengio, R.~Ducharme, P.~Vincent, and C.~Jauvin, ``A neural probabilistic
  language model,'' \emph{Journal of machine learning research}, vol.~3, no.
  Feb, pp. 1137--1155, 2003.

\bibitem{mikolov2013efficient}
T.~Mikolov, K.~Chen, G.~Corrado, and J.~Dean, ``Efficient estimation of word
  representations in vector space,'' \emph{arXiv preprint arXiv:1301.3781},
  2013.

\bibitem{mikolov2013distributed}
T.~Mikolov, I.~Sutskever, K.~Chen, G.~S. Corrado, and J.~Dean, ``Distributed
  representations of words and phrases and their compositionality,'' in
  \emph{Advances in neural information processing systems}, 2013, pp.
  3111--3119.

\bibitem{srivastava2014dropout}
N.~Srivastava, G.~Hinton, A.~Krizhevsky, I.~Sutskever, and R.~Salakhutdinov,
  ``Dropout: A simple way to prevent neural networks from overfitting,''
  \emph{The Journal of Machine Learning Research}, vol.~15, no.~1, pp.
  1929--1958, 2014.

\bibitem{villarroel2016release}
L.~Villarroel, G.~Bavota, B.~Russo, R.~Oliveto, and M.~Di~Penta, ``Release
  planning of mobile apps based on user reviews,'' in \emph{Proceedings of the
  38th International Conference on Software Engineering}.\hskip 1em plus 0.5em
  minus 0.4em\relax ACM, 2016, pp. 14--24.

\bibitem{ortu2015jira}
M.~Ortu, G.~Destefanis, B.~Adams, A.~Murgia, M.~Marchesi, and R.~Tonelli, ``The
  jira repository dataset: Understanding social aspects of software
  development,'' in \emph{Proceedings of the 11th international conference on
  predictive models and data analytics in software engineering}.\hskip 1em plus
  0.5em minus 0.4em\relax ACM, 2015, p.~1.

\bibitem{kingma2014adam}
D.~P. Kingma and J.~Ba, ``Adam: A method for stochastic optimization,''
  \emph{arXiv preprint arXiv:1412.6980}, 2014.

\bibitem{prechelt1998automatic}
L.~Prechelt, ``Automatic early stopping using cross validation: quantifying the
  criteria,'' \emph{Neural Networks}, vol.~11, no.~4, pp. 761--767, 1998.

\bibitem{fu2016differential}
W.~Fu, V.~Nair, and T.~Menzies, ``Why is differential evolution better than
  grid search for tuning defect predictors?'' \emph{arXiv preprint
  arXiv:1609.02613}, 2016.

\end{thebibliography}

\end{document}